\documentclass[pmlr,twocolumn,10pt]{jmlr} 

\mlhtrack{}

\newif\iffinal
\finalfalse  

\iffinal
    \ifmlhneedspmlr
      \jmlrvolume{XXX}
      \jmlryear{2026}
    \fi
    \ifmlhfindings \jmlrproceedings{}{ML4H 2026 - Findings Track}\fi
    \ifmlhdemo     \jmlrproceedings{}{ML4H 2026 - Demo Track}\fi
    \jmlrworkshop{Machine Learning for Health (ML4H) 2026}
\else
    \jmlrproceedings{}{Submitted to ML4H 2026: \mlhtrackname}
    \jmlrworkshop{Machine Learning for Health (ML4H) 2026}
\fi

\usepackage{booktabs}
\usepackage{siunitx}
\usepackage{longtable}
\usepackage{listings}
\usepackage{multirow}
\usepackage{placeins}
\usepackage[T1]{fontenc}

\lstdefinestyle{jsonstyle}{
    basicstyle=\ttfamily\small,
    breaklines=true,
    breakatwhitespace=false,
    columns=fullflexible,
    showstringspaces=false,
    frame=single,
    keepspaces=true,
    linewidth=\linewidth,
    aboveskip=4pt,
    belowskip=4pt
}

\newenvironment{systemprompt}{\small\begin{quote}}{\end{quote}}

\usepackage[switch]{lineno}

\theorembodyfont{\upshape}
\theoremheaderfont{\scshape}
\theorempostheader{:}
\theoremsep{\newline}

\title[Tracing the Heart]{Tracing the Heart: An Evidence-Linked Pipeline for Heart-Failure Feature Engineering}

\author{
\Name{Soorya Ram Shimgekar} \\
\addr Nimblemind \\ 
\Name{Michelle Hu} \\
\addr Nimblemind \\ 
\Name{Dorisa Shehi} \\
\addr Nimblemind \\ 
\Name{Daniel Kang} \\
\addr Nimblemind \\ 
\Name{Roy Ka-Wei Lee} \\
\addr Singapore University of Technology and Design \\
\Name{Koustuv Saha} \\
\addr University of Illinois Urbana-Champaign \\
\Name{Christian Poellabauer} \\
\addr Florida International University \\
\Name{Christopher Lee} \\
\addr University of California Los Angeles\\
\Name{Sajeev Singh} \\
\addr Department of Health Informatics, Rutgers University Newark \\
\Name{Piyum Zonooz} \\
\addr Nimblemind \\
\Name{Navin Kumar} \\
\addr Nimblemind \\
\Name{Zeeshan Ahmed} \\
\addr Rutgers Institute for Health, Health Care Policy and Aging Research \\
\addr Department of Medicine, Robert Wood Johnson Medical School, Rutgers Health \\
\Name{Priyadarshini Kachroo} \\
\addr Department of Health Informatics, Rutgers University Newark
}

\begin{document}

\maketitle

\ifmlhdemo\else
\begin{abstract}
Electronic health record (EHR) feature engineering is a major bottleneck in clinical research and AI, accounting for 39--45\% of data scientists' workload. This is especially pronounced in heart failure, which affects an estimated 6.7 million U.S. adults and requires integrating fragmented EHR data with disease-specific, guideline-based clinical reasoning. Existing rule-based and large language model (LLM)-based approaches offer only partial automation with limited maintainability and evidence traceability. We developed the Nimblemind Multi-Agent System (nMAS), an evidence-linked, rubric-grounded pipeline for automated heart-failure feature engineering, and evaluated it on 500 dummy patient records from nine EHR source tables. nMAS generated 132 structured and 70 rubric-scored aggregated features, verified for structural integrity, rubric compliance, and provenance, and audited by a restricted LLM. Adding the aggregated features improved held-out AUROC from 0.895 to 0.963 for HFrEF and 0.870 to 0.910 for HFpEF phenotyping, and an independent LLM-based rubric assessment of evidence support and methodological soundness scored the features at 81.5\% of maximum points. These results demonstrate the feasibility of automated, auditable feature engineering for complex cardiovascular EHR data, though evaluation was limited to a single-institution cohort and external validation is needed.
\end{abstract}
\begin{keywords}
electronic health records, heart failure, feature engineering, clinical artificial intelligence, multi-agent systems, evidence-linked extraction
\end{keywords}
\fi

\ifmlhneedsstatements
\paragraph*{Data and Code Availability}
This initial paragraph is \textbf{mandatory}. Briefly state what data you
use (including citations if appropriate) and whether and where the data are
available to other researchers.
If you are not sharing code, you must explicitly state that you are not
making your code available. If you are making your code available, then
at the time of submission for review, please include your code as
supplemental material or as a code repository link; in either case, your
code must be anonymized. If your paper is accepted, then you should
de-anonymize your code for the camera-ready version of the paper. \emph{If
you do not include this data and code availability statement for your
paper, or you provide code that is not anonymized at the time of
submission, then your paper will be desk-rejected.} Your experiments later
could refer to this initial data and code availability statement if it is
helpful (e.g., to avoid restating what data you use).

\paragraph*{Institutional Review Board (IRB)}
This initial paragraph is \textbf{mandatory}. If your research requires IRB
approval or has been designated by your IRB as Not Human Subject
Research, then for the camera-ready version of the paper, you must
provide IRB information (and at the time of submission for review, you
can say that this IRB information will be provided if the paper is
accepted). If your research does not require IRB approval, then you
must state this to be the case.
\fi

\section{Introduction}
\label{sec:intro}

Before a clinical machine-learning model can be trained or a population-level cohort study can be conducted, routinely collected clinical data must be transformed into reliable, analysis-ready patient-level variables. This feature-engineering process involves extracting, cleaning, harmonizing, and deriving variables from heterogeneous clinical records and represents a substantial component of healthcare data-science workflows. Data scientists report spending approximately 39--45\% of their working time on data loading and cleaning rather than analysis \citep{anaconda2020datascience}. At the median U.S. data-scientist wage of \$112{,}590 per year, or approximately \$54 per hour \citep{bls2026datascientists}, this corresponds to roughly \$845--\$975 of labor per 40-hour workweek devoted to data preparation. These costs are further compounded when feature-engineering workflows must be repeatedly reconstructed for new datasets, institutions, or research questions. The resulting lack of reusable data infrastructure remains a barrier to the broader deployment of artificial intelligence (AI) in healthcare \citep{bian2026learningutility}.

Cardiology presents a particularly demanding setting for clinical feature engineering because clinically meaningful variables are distributed across multiple EHR domains and often require longitudinal interpretation. Diagnoses, medications, laboratory measurements, procedures, and imaging findings may be stored in separate relational tables rather than in a single analysis-ready dataset \citep{kao2022ehrhf}. Consequently, constructing patient-level features requires more than extracting individual variables: heterogeneous records must be reconciled, temporally aggregated, and transformed according to clinically meaningful definitions \citep{roger2021hfepidemiology}. Heart failure (HF) provides a representative case because its characterization depends on multiple complementary sources of clinical evidence and affects a large and growing patient population. An estimated 6.7 million U.S. adults have HF, with prevalence projected to increase to 8.7 million by 2030 \citep{bozkurt2025hfstats}. In 2023, HF was mentioned on 452{,}573 U.S. death certificates, corresponding to 14.6\% of all deaths \citep{nchs2025mcdwonder}.

HF phenotyping also requires the application of explicit clinical definitions to heterogeneous measurements. Contemporary classifications use guideline-specified thresholds, such as ejection-fraction categories, while other features depend on laboratory measurements, medication histories, comorbidities, and disease-specific scoring criteria \citep{ostrominski2024hfguidelines,maddox2024hfref}. These requirements create a distinction between simply extracting information from an EHR and constructing a clinically meaningful feature: the latter requires explicit rules describing which source variables should be used, how they should be combined, and how the resulting value should be interpreted. For research applications, these transformations must additionally be reproducible and traceable to their underlying source data.

These requirements motivate an automated feature-engineering approach that combines heterogeneous EHR integration with explicit clinical specifications and evidence traceability. We address this need with the Nimblemind Multi-Agent System (nMAS) \citep{shimgekar2025agentic}, an evidence-linked, rubric-grounded pipeline for heart-failure feature engineering. Rather than generating unrestricted clinical summaries, nMAS applies deterministic scoring rules derived from clinical specifications and uses a restricted language-model auditing step to evaluate the resulting feature construction. Each derived feature remains linked to its source EHR variables and predefined scoring criteria, enabling automated processing while preserving interpretability and auditability.

In this pilot study, we applied nMAS to 500 dummy heart-failure patient records assembled from nine EHR source tables. We evaluated the resulting dataset for structural integrity, rubric compliance, and clinical coherence. Although downstream predictive performance is reported as a secondary evaluation, the primary objective is to assess whether automated feature engineering produces structurally valid, rubric-compliant, and clinically coherent patient-level features. This study therefore provides preliminary evidence for the feasibility of evidence-linked, rubric-grounded feature generation as a reusable approach for preparing complex cardiovascular EHR datasets.

\section{Related Work}
\label{sec:related_work}

Artificial intelligence has been widely applied to diagnosis, risk stratification, and disease management, but most systems operate on a single data modality rather than integrating longitudinal EHR data \citep{boccuto2023aihf}. Many derive structured clinical variables from a single rich source, such as unstructured clinical documents processed with large language models \citep{truhn2024gpt4pathology,balasubramanian2025llmpathology}, instead of reconciling heterogeneous data across multiple EHR domains. The same pattern exists in cardiology, where AI models have been developed for electrocardiogram-based heart-failure risk prediction \citep{dhingra2025aiecg}. However, these single-modality approaches do not address the preprocessing and integration of structured EHR data required for heart-failure phenotyping, which depends on combining guideline-defined ejection-fraction categories, medication history, laboratory measurements, comorbidities, and other clinical variables across multiple EHR domains \citep{roger2021hfepidemiology}.

General EHR feature-engineering frameworks address part of this preprocessing challenge but remain disease agnostic. FIDDLE and MIMIC-Extract automate preprocessing and temporal aggregation to generate machine learning-ready representations \citep{tang2020fiddle,wang2020mimicextract}; the OHDSI patient-level prediction framework derives standardized covariates from common clinical domains \citep{reps2018plp}; and rule-based methods such as the Charlson and Elixhauser indices engineer comorbidity features from diagnosis codes \citep{quan2005comorbidity}. More recent LLM-based approaches relax the rule-based constraint but not the need for guideline-grounded reasoning: FeatEHR-LLM generates feature-extraction code from dataset schemas rather than clinical literature \citep{karami2026featehrllm}, while AgentScore learns candidate scoring rules validated statistically against patient data rather than grounded in clinical evidence \citep{ruhrberg2026agentscore}. Consequently, these approaches cannot derive many clinically meaningful heart-failure predictors, including phenotype, disease severity, guideline-directed medical therapy, and care gaps, which require synthesizing diagnoses, medications, laboratory results, imaging, procedures, and behavioral history using guideline-based clinical reasoning.

Few reusable frameworks therefore address the full feature-engineering requirements of cardiovascular research. Heart-failure phenotyping requires integrating diagnoses, medications, laboratory results, imaging, procedures, behavioral history, and guideline-defined ejection-fraction thresholds across longitudinal EHR data. Because these derived variables govern clinical risk stratification, cohort eligibility, and treatment assessment, feature engineering requires both guideline-based clinical reasoning and an auditable link between every derived feature and its supporting evidence.

The present work addresses this gap with an evidence-linked, rubric-grounded pipeline for automated heart-failure feature engineering. Unlike existing frameworks that emphasize generic preprocessing or data-driven rule discovery, our approach encodes guideline-based clinical reasoning within an explicit rubric, integrates heterogeneous EHR evidence into clinically meaningful features, and preserves traceability from each feature to its underlying evidence and source records.

\section{Data}
\label{sec:data_task}
The evaluation dataset consisted of 500 dummy patient records drawn from an aggregated cohort modeled on REDACTED, a cardiovascular program serving a minoritized population \citep{pina2021racehf}. Patients were included if they had a documented heart-failure diagnosis in the source EHR; no additional
inclusion or exclusion criteria were applied. Cohort characteristics are summarized in Table~\ref{tab:cohort}. Each patient was represented across nine structured EHR tables covering demographics, other diagnoses, medication orders, echocardiogram procedures, surgical procedures, laboratory components and external labs, ejection fraction flowsheet entries, and social history (Table~\ref{tab:ehr_data} in the Appendix). Direct patient identifiers were replaced with de-identified identifiers before analysis, while clinically relevant information required for feature extraction and phenotyping was preserved. 

Each record combined coded administrative and demographic fields (age, sex, race, ethnicity, language, vitals) with free-text clinical fields requiring pattern-based extraction, including ICD-10 code lists, diagnosis names, medication display and pharmacologic-class names, procedure display names, ejection fraction flowsheet text, and social history narrative.

\section{Method}
\label{sec:method}

\begin{figure}[htbp]
\floatconts
{fig:architecture}
{\caption{Overview of the nMAS heart-failure feature-extraction pipeline.}}
{\includegraphics[width=\linewidth]{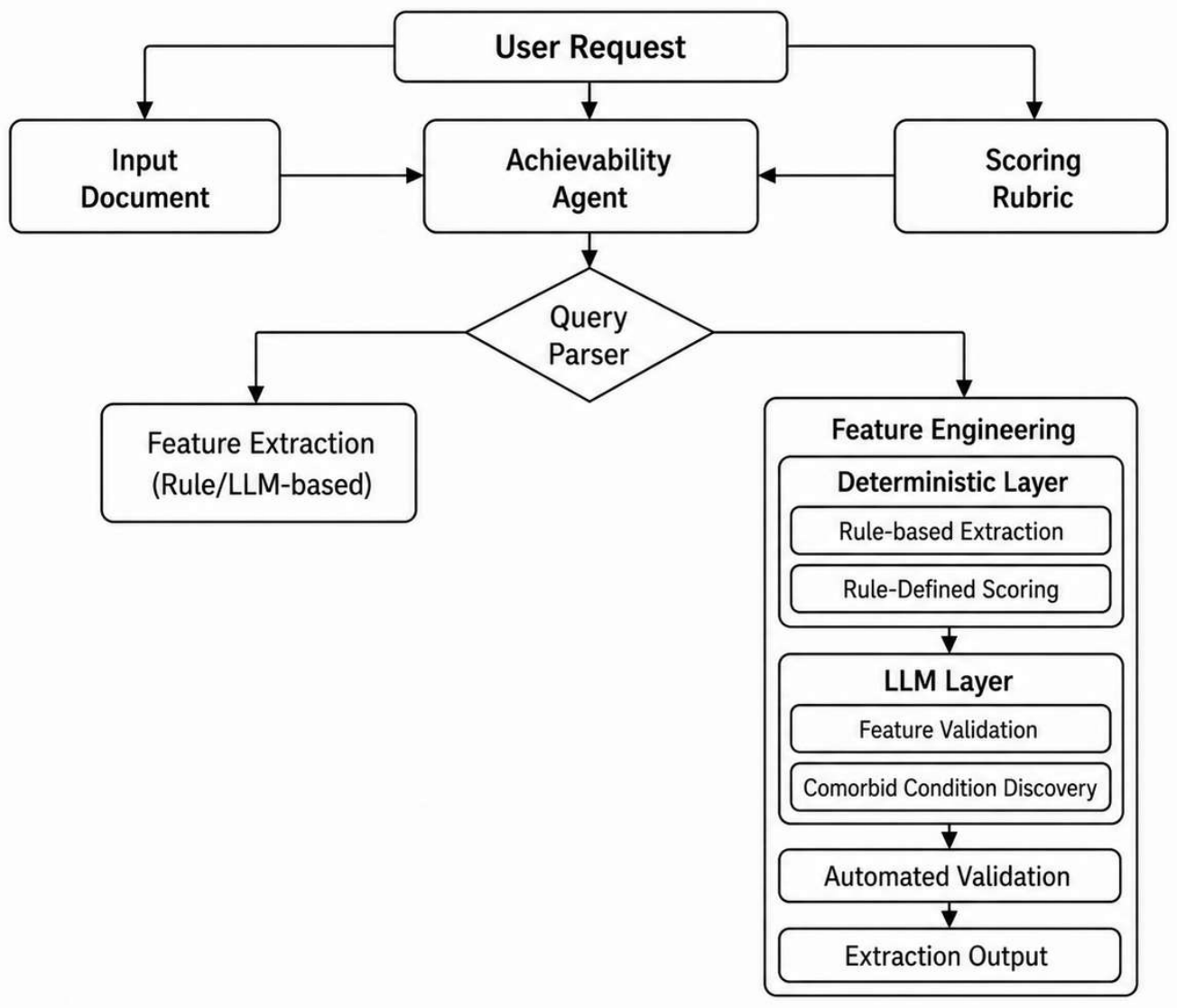}}
\end{figure}

\textbf{Workflow Overview.}
\label{sec:workflow_overview}
As shown in Figure~\ref{fig:architecture}, the workflow begins with a request specifying the input EHR exports and constructs the target features defined in the versioned rubric. The Achievability Agent first checks that the available source data can support each target feature; features whose required inputs are absent are reported as unsupported rather than assigned inferred values. For supported requests, the Query Parser then routes the request to one of several nMAS pipelines \citep{shimgekar2025agentic}; for brevity, we provide two representative pipelines here: feature extraction and feature engineering. 

In the feature-extraction pipeline, each field is routed through rule/LLM-based extraction tiers, depending on how much contextual interpretation the field requires, and returns the field's value directly from the source document along with a verbatim source sentence confirming that value \citep{wang2026hpylori}. Feature engineering instead composes extracted values into rubric-scored, evidence-linked composite variables that remain traceable to their source data and to the guideline literature that justifies them. This paper focuses exclusively on the feature-engineering pipeline, in which supported features are produced in two stages, with routing driven by the configured rubric
and field definitions rather than by any downstream label.
Stage~1 builds a single patient-level table from the EHR
source tables: (1) standardizes each table (whitespace, null tokens, uppercased identifiers, and ISO datetimes); (2) deduplicates each table on a clinically meaningful event key; (3) temporally aggregates each source to the patient level independently; (4) Merges  per-table summaries into unique patient rows.

Stage~2 enriches this table with higher-order composite features: (1) produces the higher-order feature categories described below, including a
comorbid-burden category spanning five organ systems, each returned as a compact value paired with a structured evidence trace; (2) reviews composite features with an LLM auditor; (3) checks structural integrity, rubric compliance, monotonicity, and evidence traceability before release.

\textbf{Input Standardization and Deduplication.}
\label{sec:standardization}
Before aggregation, every source table was converted into a consistent representation. String columns were whitespace-collapsed and trimmed, empty strings and literal \texttt{"nan"} tokens were converted to null, and patient identifiers were uppercased so that keys join reliably. EHR-style datetime fields (for example, \texttt{03-Apr-24 06.00.00.000000 AM}) were parsed into a unified ISO representation (\texttt{2024-04-03 06:00:00}), which is required for correct temporal ordering, first/latest event extraction, and recurrence analysis.

Each table was deduplicated using clinically meaningful event keys to prevent repeated exports or re-orders from inflating counts. The source tables and keys are listed in Appendix Table~\ref{tab:sources}. Deduplication preceded aggregation so that, for example, re-sent medications and retransmitted laboratory results were not double counted.

\textbf{Patient-Level Temporal Aggregation.} Each source was collapsed to the patient level first, computing counts, first and latest event timestamps, and clinically relevant flags. Table~\ref{tab:aggregates} in the Appendix lists the quantities produced from each event source.

Ejection fraction is often recorded as a range \citep{ostrominski2024hfguidelines}, so each entry was parsed into a low bound, a high bound, and a midpoint (``20--25\%'' becomes 20, 25,
and 22.5), from which the lowest, highest, and most-recent EF were computed, plus a flag for latest recordings with conflicting values. On merging, every continuous measurement (EF, BNP, troponin, BMI, blood pressure) received an availability indicator, and missing values were kept missing rather than set to zero, since a zero EF would read as profound systolic dysfunction and inflate the severity scores. The merge was validated to keep one row per patient and preserve the input row count.

\textbf{Clinical Feature Engineering.} Before scoring, a clinical-reasoning layer converted raw text and codes into interpretable flags (Table~\ref{tab:signals} in the Appendix). Two choices are worth noting. First, behavioral exposures were kept as ordered categories rather than binary flags, because cardiovascular risk from tobacco and alcohol persists after cessation and varies by intensity \citep{yoo2023smoking,piano2025alcohol}. Second, the heart-failure phenotype was dictated by ICD-derived values rather than diagnosis names, carrying a conflict flag when the two disagreed \citep{quan2005comorbidity}.

\textbf{Composite Feature Generation.}
In Stage 2, the pipeline generates composite features by applying a versioned clinical scoring rubric to Stage 1 variables. These features summarize multiple clinical inputs into interpretable scores, counts, and indices while preserving their contributing evidence. The rubric defines the conditions, point assignments, and evidence hierarchy used to transform structured EHR data into interpretable feature scores.

Table \ref{tab:full_rubric} in the Appendix provides the complete point
structure for all eight feature categories, including
each component's condition and point value. A representative portion of the
underlying JSON representation is shown in Figure~\ref{fig:rubric} in the Appendix.

The rubric was developed using \texttt{Qwen 2.5-1.5B-Instruct},
which assembled recent guideline-level literature into a
proposed scoring framework. A professional cardiologist
then reviewed and approved the rubric in full. The final
version included 22 references from 2023 onward, sourced
from leading cardiovascular and specialty journals, with
each reference linked to the component it supports.

The scoring system preserves clinically meaningful ordering
across structured variables. The weights distinguish: (1) direct evidence from supportive evidence; (2) current findings from historical information; (3) independent features from overlapping measurements; and (4) feature-specific variables from cross-system modifiers. The assigned point values represent an interpretable
engineering design informed by clinical literature rather
than quantities directly derived from published studies.

In general, direct diagnostic evidence receives the highest weight within each category. Supportive laboratory findings and cross-system modifiers contribute smaller, capped amounts. For example, hypertension diagnosis contributes fewer cardiovascular risk points because measured blood pressure is scored separately. Elevated natriuretic peptide and troponin increase disease severity because they reflect worsening cardiac structure and function \citep{myhre2024biomarkers}. Current tobacco use receives greater weight than former use because smoking cessation reduces heart failure risk \citep{yoo2023smoking}. Alcohol use is ordered by consumption intensity following American Heart Association guidance \citep{piano2025alcohol}. Age and blood pressure weights follow contemporary cardiovascular risk equations validated across diverse populations \citep{lewis2025prevent}. Demographic vulnerability variables are included only as administrative equity review factors rather than biological risk measures \citep{bazoukis2025sdoh}.

After validation, patient data are processed through the
deterministic scoring engine using the finalized rubric.
Seven of the eight category features use the same scoring
approach. These include disease severity, cardiovascular
risk, the five comorbid-burden scores (kidney, lungs,
metabolic, blood, and brain-health context), and demographic
vulnerability. Each feature sums component points, caps the total at 100,
and maps the score to an ordinal label. Scores of 55 or
above are classified as high, 25 to 54 as medium, and
below 25 as low. Each feature also returns the contributing components,
assigned points, plain-language explanations, source
columns, and whether required inputs were present. This
ensures that every aggregated value remains auditable.

The heart-failure phenotype feature uses a separate logic
because ejection-fraction classes are categorical thresholds
rather than weighted composites. The phenotype is assigned using a precedence rule over available evidence. When numeric EF is present, it directly determines the phenotype: heart failure with reduced ejection fraction (HFrEF) at midpoint $\leq 40$, heart failure with mildly reduced ejection fraction (HFmrEF) below 50, and heart failure with preserved ejection fraction (HFpEF) otherwise, following contemporary guideline EF categories \citep{ostrominski2024hfguidelines}. When EF is unavailable, the pipeline uses ICD-10 diagnosis codes for systolic, diastolic, or combined heart failure without imputing an EF value. A patient is never assigned both reduced and preserved phenotypes. This same numeric-or-diagnosis-only logic determines eligibility for HFrEF-specific care-gap checks, keeping guideline-directed therapy recommendations aligned with available evidence \citep{maddox2024hfref}.

\textbf{LLM Rubric-Auditor Validation and Correction.} Because composite features encode clinical reasoning that deterministic rules may approximate imperfectly, each candidate row is audited by \texttt{Qwen 2.5-1.5B-Instruct}. The prompt provides patient evidence, the candidate row, and a whitelist of correctable fields; the model returns \texttt{\{"valid": true\}} when the row is correct or a minimal \texttt{corrections} object otherwise. Only whitelisted numeric and categorical fields may be modified (Table~\ref{tab:llm_fields} in the Appendix); fields ending in \texttt{\_json} or \texttt{\_explanation} are protected to preserve deterministic evidence traces and rationales. Corrections are type-coerced before application, while unparseable responses are logged as audit failures and leave the original scores unchanged. The prompt template is provided in Figure~\ref{fig:auditor_prompt} in the Appendix.

The auditor runs locally, and each row records the model, raw response, and \texttt{generated\_*} status, ensuring that auditing is a required and traceable pipeline step rather than an optional post-processing stage.

\textbf{Comorbid Condition Discovery.} A final stage catalogs conditions beyond heart failure that the dataset can model. Candidate conditions are fixed in the rubric (13 cardiovascular and comorbid conditions, each tagged by clinical group and evidence type), so the stage selects supporting evidence rather than discovering new conditions. For each candidate, the pipeline considers only merged and higher-order columns in the final dataset, excludes conditions without supporting columns, and prompts the language model, under the auditor's guardrails, to identify evidence-bearing columns and explain their relevance. The model cannot add conditions, invent columns, or label patients; returned columns are validated against the dataset, with a deterministic fallback for unusable responses. The resulting condition-prediction catalog and prompt template are provided in Figure~\ref{fig:condition_prompt} in the Appendix.

\textbf{Quality Assurance and Evaluation.}
The enriched dataset was validated using an automated evaluation harness that performs structural, clinical, and provenance checks and generates a machine-readable report of identified issues. The harness verifies: (1) unique patient identifiers and valid table joins; (2) valid and mutually exclusive heart-failure phenotype indicators; (3) preservation of missing values; (4) compliance with the scoring rubric by independently checking selected scores against predefined thresholds and formulas defined in Figure~\ref{fig:rubric}; (5) expected scoring behavior, such as higher scores for current versus former tobacco use; (6) complete traceability of score components; and (7) provenance of generated features and supporting references.

\textbf{Multi-Agent System Evaluation.}
To evaluate whether aggregated features improve prediction beyond cleaned EHR variables alone, we trained XGBoost classifiers \citep{chen2016xgboost} for two binary tasks: HFrEF (reduced ejection fraction, EF $\leq 40$) versus phenotype-unknown and HFpEF (preserved ejection fraction, EF $\geq 50$) versus phenotype-unknown. As described in Section~\ref{sec:data_task}, this cardiovascular program serves a minoritized population where limited access to diagnostic imaging and specialist follow-up can leave heart failure documented without definitive phenotype assignment \citep{pina2021racehf}. Consequently, a representative non-heart-failure control cohort was unavailable for this pilot study. Phenotype-unknown patients had documented heart failure but insufficient evidence to assign HFrEF, HFmrEF, or HFpEF; thus, the evaluation assessed heart-failure phenotyping rather than disease diagnosis.

For each task, we compared a baseline containing only cleaned and merged structured variables with an aggregated feature set that additionally included the composite features. XGBoost was chosen because it natively handles missing values during split selection, eliminating explicit imputation. All models used identical hyperparameters (Appendix~\ref{app:hyperparameters}).

Models were evaluated using 5-fold stratified cross-validation repeated 10 times with independent shuffles (50 test evaluations per model). The same fold assignments were reused for both feature sets within each task to enable paired comparisons on identical patients. We report mean $\pm$ standard deviation for accuracy, AUROC, and F1 score across test folds. Feature contributions were quantified using global TreeSHAP values \citep{lundberg2017shap}.

\textit{LLM-Based Evaluation.} To assess feature quality beyond predictive performance, we performed a structured construct-validity evaluation of every engineered rubric component using Claude Opus 4.8 and a five-dimensional rubric. The evaluation assessed external evidence support, methodological soundness, reproducibility, and EHR data reliability, complementing downstream classification results. The complete rubric, procedure, system prompt, metadata, and results are provided in Appendix~\ref{app:llm_construct_validity}.

\textbf{Comparator and ablation studies.} We conducted two additional analyses: an ablation study measuring the contribution of individual feature categories and rubric design choices, and an independent comparator where an LLM engineered features directly from raw EHR data. Full methodologies, results, and prompts are provided in Appendix~\ref{app:ablation} and Appendix~\ref{app:claude_prompt}.

\section{Results}
\label{sec:results}
\textbf{Structured Merge.}
Stage 1 collapsed nine source tables to the patient level without Cartesian explosion, producing exactly one row per patient, no duplicates, and 132 structured columns across all 500 patients. The engineered features showed coherent cohort-level distributions (Table~\ref{tab:cohort}): cardiovascular comorbidities were common, while key measurements remained sparse. Notably, numeric ejection fraction was available for only 3.0\% of patients, requiring diagnosis-name fallback for phenotype assignment.

\begin{table}[htbp]
\floatconts
  {tab:cohort}%
  {\caption{Characteristics of the 500-patient development
cohort. Values shown are dummy data for illustration.}}%
  {\footnotesize
  \setlength{\tabcolsep}{8pt}
  \renewcommand{\arraystretch}{1.15}
  \begin{tabular}{@{}lr@{}}
  \toprule
  \textbf{Characteristic} & \textbf{Value} \\
  \midrule
  Total patients & 500 \\
  \addlinespace
  \multicolumn{2}{@{}l}{\textit{Sex}} \\
  \quad Male & 256 (51.2\%) \\
  \quad Female & 243 (48.6\%) \\
  \quad Unspecified & 1 (0.2\%) \\
  \addlinespace
  Mean BMI (n${=}$266) & 29.99 \\
  \addlinespace
  \multicolumn{2}{@{}l}{\textit{Documented comorbidities}} \\
  \quad Chronic kidney disease & 227 (45.4\%) \\
  \quad Diabetes & 208 (41.6\%) \\
  \quad Coronary artery disease & 202 (40.4\%) \\
  \quad Atrial fibrillation & 175 (35.0\%) \\
  \addlinespace
  \multicolumn{2}{@{}l}{\textit{Measurement availability}} \\
  \quad Numeric ejection fraction & 15 (3.0\%) \\
  \quad BNP / NT-proBNP & 154 (30.8\%) \\
  \quad Troponin & 429 (85.8\%) \\
  \bottomrule
  \end{tabular}}
\end{table}

\textbf{Higher-Order Composite Features.}
Stage 2 generated the feature categories in Table~\ref{tab:full_rubric} as 70 columns, with all 500 rows audited by \texttt{Qwen 2.5-1.5B-Instruct}. Each feature included a structured evidence trace documenting contributing components, point values, source columns, and input availability. The categories produced two output types, illustrated by representative examples.

The first was a scored assessment with ranked drivers, shown in Figure~\ref{fig:disease_severity} using the disease-severity feature. Because this patient lacked numeric ejection fraction, phenotype was inferred from diagnosis text (\texttt{HFpEF\_by\_diagnosis\_name}, source \texttt{DXNAMES}). The resulting score was driven by elevated natriuretic peptide levels, abnormal troponin results, and chronic kidney disease, each recorded as a source-linked contributor.

The second output type was a documentation-gap list, shown in Figure~\ref{fig:care_recommendation}. The care-recommendation feature identifies evidence columns associated with each gap and frames outputs as documentation review rather than prescriptive advice. For example, a patient with coronary artery disease but no statin flag received a high-urgency statin documentation review prompt, along with prompts to review potassium, creatinine, and sodium monitoring.

\textbf{Comorbid Condition Discovery.}
The condition-discovery stage indicates which other conditions the
enriched dataset could support for modeling. It produced a catalog of
thirteen candidate conditions spanning cardiac, vascular, metabolic, renal,
pulmonary, and hematologic domains, for example hypertension, diabetes,
chronic kidney disease, coronary artery disease, atrial fibrillation,
anemia, and COPD. For each, it listed the merged and higher-order columns
present in the final dataset that could serve as modeling evidence, together
with a generated rationale and no patient-level diagnoses. The anemia entry
in the Appendix (Figure~\ref{fig:comorbid_condition_example}) is representative: an anemia diagnosis flag, hemoglobin laboratory
summaries, and blood and kidney comorbid-burden features together support
modeling that condition.

\textbf{Evaluation Performance.}  Table~\ref{tab:classification_results} summarizes the 5-fold cross-validated performance of the baseline and aggregated feature sets. Adding the composite features improved performance for both classification tasks across all evaluation metrics. The largest improvement was observed for HFrEF classification, with accuracy increasing from 0.776 to 0.896 (+0.120). HFpEF classification also improved across all evaluated metrics, although the gains were smaller, with accuracy increasing from 0.752 to 0.809 (+0.057).

\begin{table*}[htbp]
\floatconts
  {tab:classification_results}%
  {\caption{Held-out cross-validated performance (mean $\pm$ std across 5
folds, repeated 10 times) for baseline versus aggregated feature sets, for each phenotype
classification task.}}%
  {\footnotesize
  \setlength{\tabcolsep}{3pt}
  \begin{tabular}{@{}llccc@{}}
  \toprule
  \textbf{Task} & \textbf{Features} & \textbf{Accuracy} & \textbf{AUROC} & \textbf{F1} \\
  \midrule
  \multirow{2}{*}{HFrEF} & Baseline   & 0.776 $\pm$ 0.065 & 0.895 $\pm$ 0.046 & 0.776 $\pm$ 0.063 \\
   & Aggregated & 0.896 $\pm$ 0.049 & 0.963 $\pm$ 0.028 & 0.894 $\pm$ 0.052 \\
  \midrule
  \multirow{2}{*}{HFpEF} & Baseline   & 0.752 $\pm$ 0.056 & 0.870 $\pm$ 0.042 & 0.758 $\pm$ 0.058 \\
   & Aggregated & 0.809 $\pm$ 0.061 & 0.910 $\pm$ 0.040 & 0.812 $\pm$ 0.059 \\
  \bottomrule
  \end{tabular}}
\end{table*}
Global feature importance was assessed using mean absolute SHAP values. For HFrEF classification, 6 of the top 10 features were composite features generated by the proposed framework, with demographic vulnerability adjusted gap and disease severity adjusted gap ranking above the highest-ranked baseline feature, HF GDMT medication class count. For HFpEF, two composite features appeared in the top 10, with disease severity score as the most influential composite predictor and troponin abnormal count remaining the dominant baseline feature. Complete SHAP rankings are provided in Table~\ref{tab:shap_summary} in the Appendix.

The LLM-based construct-validity assessment showed strong overall performance, with most categories exceeding 85\% and an overall normalized score of 81.5\% (Table~\ref{tab:llm_construct_validity_results}). Lower scores were observed for demographic vulnerability and gap features, with care recommendations receiving the lowest score (38.5

To determine whether performance gains reflected clinically grounded feature weighting rather than feature count alone, equal-weight and random-weight ablations fixed the aggregated feature set while varying only the scoring logic. Both substantially reduced performance (Appendix~\ref{app:ablation}), indicating that rubric weighting contributed to the improvement. Full ablation and comparator analyses are reported in Appendix~\ref{app:ablation} and Appendix~\ref{app:claude_prompt}.

\section{Discussion}
\label{sec:discussion}
We developed a pipeline that transforms fragmented EHR exports into a patient-level heart-failure cohort with rubric-derived, evidence-linked feature categories. Adding these features improved HFrEF accuracy, AUROC, and F1 score by 0.120, 0.068, and 0.118, respectively, and HFpEF performance by 0.057, 0.040, and 0.054, respectively. Independent literature-grounded grading scored seven of eleven feature categories above 85/100. Feature-importance analysis showed that rubric-derived features accounted for 6 of the top 10 HFrEF predictors and 2 of the top 10 HFpEF predictors, suggesting they captured information beyond the original structured variables.

Although not designed for clinical decision support, the generated features may support identification of patients relevant to established care pathways. For example, care-gap checks identify reduced-EF heart-failure patients lacking guideline-directed medical therapy, a population relevant to optimization strategies such as STRONG-HF \citep{mebazaa2022stronghf}, while kidney comorbidity features may help identify patients who could benefit from earlier nephrology referral, which has been associated with improved outcomes in observational studies and meta-analyses \citep{cheng2025ckdreferral}. These represent potential applications rather than measured clinical effects.

Beyond predictive performance, the pipeline provides a reproducible and auditable approach for clinical data preparation. The generated features retain links to their source variables, scoring criteria, and supporting evidence, enabling transparent review of how each feature was constructed. This design separates clinical knowledge representation from feature execution, allowing the scoring framework to be updated independently from the underlying data-processing pipeline. The same separation makes the approach reusable well beyond heart failure. Any data-intensive chronic disease facing the same feature-engineering bottleneck could substitute its own guideline-grounded rubric into this scoring engine without rebuilding the extraction and auditing infrastructure underneath it. The condition-discovery catalog is one concrete instance of this reusability, extending the framework across thirteen cardiovascular conditions and allowing future studies to reuse feature-generation methods rather than repeat manual preparation.

Constructing analysis-ready cohorts from fragmented EHR exports remains a major bottleneck in clinical research. As noted in Section~\ref{sec:intro}, data preparation accounts for a substantial portion of data scientists’ time \citep{anaconda2020datascience}, making reusable, version-controlled pipelines valuable for reducing repetitive manual work. This burden extends to clinical workflows: in a heart-failure clinical trial prescreening effort, manual eligibility review took approximately 15--20 minutes per patient, with screening of 198 patients requiring two weeks of coordinator effort \citep{adupa2016hfpefprescreen}.

The pipeline combines deterministic, guideline-grounded scoring with bounded model-assisted rubric generation and auditing. Rule-based approaches provide transparency but can become brittle as schemas evolve \citep{wang2018clinicalie}, whereas large language models offer flexible
clinical abstraction but may produce unsupported outputs \citep{ji2023hallucination}. By combining structured execution with evidence-linked rubric development, this framework extends prior evidence-linked clinical AI approaches \citep{shimgekar2025agentic,shimgekar2025nimblelabs,vassef2025onevlm,
wang2026hpylori} toward reusable EHR feature engineering.

Several limitations should be considered. The pipeline was evaluated on a dummy, single-institution cohort of 500 patients, limiting statistical power and external generalizability. HFrEF and HFpEF were classified against phenotype-unknown patients rather than non-HF controls, so performance reflects phenotype characterization within an HF cohort rather than general disease detection. Evaluation focused on internal consistency, feature traceability, and downstream classification rather than external clinical validity, and some aggregated features incorporate phenotype-related evidence, creating potential feature-label overlap in performance and ablation analyses. Prospective clinical impact and patient outcomes were not assessed. Comorbidity and behavioral features rely partly on regular-expression matching and may miss atypical or institution-specific language. Finally, clinician review covered the extraction rubric and representative outputs rather than the full cohort.

Future work should evaluate the pipeline on larger, multi-institutional cohorts with temporally separated validation sets to assess generalizability. Broader evaluation should include both heart-failure and non-heart-failure populations, independent clinician-adjudicated phenotypes, and additional phenotyping and outcome-prediction tasks, as well as prospective workflow impact \citep{collins2024tripodai}. Until such validation, the pipeline should remain an evidence-linked feature-engineering layer for research and clinical review rather than autonomous clinical decision-making.

\bibliography{ref}

\newpage
\appendix

\onecolumn
\section{Supplementary Material}
This appendix provides supplementary material referenced throughout the main text but omitted for space. \sectionref{apd:first} gives the complete point-based scoring rubric underlying all composite features. The remaining subsections provide implementation and evaluation details: EHR source-table schema and preprocessing, model and prompt configurations for the rubric auditor and comorbid-condition discovery, the full LLM construct-validity rubric and results, composite-feature output examples, SHAP feature-importance rankings, the Claude comparator prompt, and the ablation study details and results referenced in Sections~\ref{sec:method} and~\ref{sec:results}.

\subsection{Complete Feature Scoring Rubric}\label{apd:first}
\begingroup
\setlength{\tabcolsep}{4pt}
\renewcommand{\arraystretch}{1}
\setlength{\LTcapwidth}{\linewidth}
\begin{longtable}{@{}
>{\footnotesize\raggedright\arraybackslash}p{0.18\linewidth}
>{\footnotesize\raggedright\arraybackslash}p{0.20\linewidth}
>{\footnotesize\raggedright\arraybackslash}p{0.41\linewidth}
>{\footnotesize\raggedright\arraybackslash}p{0.13\linewidth}@{}}
\toprule
\textbf{Feature} & \textbf{Component} & \textbf{Condition} & \textbf{Points} \\
\midrule
\endfirsthead
\multicolumn{4}{c}{\footnotesize\tablename\ \thetable{} -- continued from previous page} \\
\toprule
\textbf{Feature} & \textbf{Component} & \textbf{Condition} & \textbf{Points} \\
\midrule
\endhead
\midrule
\multicolumn{4}{r}{\footnotesize continued on  next page} \\
\endfoot
\endlastfoot
Disease severity & EF midpoint & $<30$ (severely reduced) & 35 \\
 & & 30--40 (reduced) & 25 \\
 & & 40--50 (mildly reduced) & 10 \\
 & & Missing EF (inferred from DX) & 8 \\
 & Biomarkers & BNP max $\geq 900$ & 20 \\
 & & $300 \leq$ BNP max $< 900$ & 10 \\
 & Troponin & per abnormal result & 3 (max 24) \\
 & Procedures & per surgery/procedure & 2 (max 12) \\
 & & LVAD implantation & 18 \\
 & Comorbidity & chronic kidney disease (DX) & 6 \\
\midrule
Cardiovascular risk & Age (cont.) & 35 (min) to 95 (max) & max 45 \\
 & Systolic BP (cont.) & 110 (min) to 180 (max) & max 18 \\
 & Diastolic BP (cont.) & 70 (min) to 110 (max) & max 8 \\
 & Risk factor & coronary artery disease (DX) & 10 \\
 & & diabetes (DX) & 8 \\
 & & chronic kidney disease (DX) & 8 \\
 & & hypertension (DX) & 6 \\
 & & hyperlipidemia (DX) & 5 \\
 & & tobacco, current & 8 \\
 & & tobacco, former & 3 \\
 & & alcohol, daily & 3 \\
\midrule
Kidney burden & Primary indicator & chronic kidney disease (DX) & 35 \\
 & & acute kidney failure (DX) & 25 \\
 & Lab modifier & creatinine abnormalities & 4 (max 20) \\
 & & latest creatinine $>1.3$ & 5 \\
 & & latest creatinine $>2.0$ & 10 \\
 & & electrolyte (Na/K) abnormalities & 2 (max 10/type) \\
 & Contextual & diabetes or hypertension (DX) & 5 each \\
\midrule
Lungs burden & Primary indicator & COPD (DX) & 35 \\
 & & sleep apnea (DX) & 25 \\
 & Risk context & tobacco, current & 12 \\
 & & tobacco, former & 5 \\
 & & BMI $\geq 35$ & 8 \\
 & & BMI 30--34.9 & 4 \\
\midrule
Metabolic burden & Diagnosis flag & diabetes (DX) & 25 \\
 & & obesity (DX) & 20 \\
 & & hyperlipidemia (DX) & 10 \\
 & & hypertension (DX) & 8 \\
 & Physical/lab marker & BMI $\geq 35$ & 20 \\
 & & BMI 30--34.9 & 12 \\
 & & glucose abnormalities & 3 (max 15) \\
 & & latest glucose $>126$ & 5 \\
 & & latest glucose $>200$ & 10 \\
\midrule
Blood burden & Primary indicator & anemia (DX) & 35 \\
 & Lab modifier & hemoglobin abnormalities & 5 (max 25) \\
 & & hemoglobin $<12$ & 10 \\
 & & hemoglobin $<10$ & 20 \\
 & Contextual & chronic kidney disease modifier & 5 \\
\midrule
Brain-health risk & Scoring factor & age $\geq 75$ & 8 \\
 & & hypertension (DX) & 12 \\
 & & diabetes (DX) & 10 \\
 & & hyperlipidemia (DX) & 8 \\
 & & sleep apnea (DX) & 5 \\
 & & tobacco, current & 8 \\
 & & alcohol, daily & 5 \\
\midrule
Demographic vulnerability & Language & non-English language & 35 \\
 & Ethnicity & Hispanic/Latino & 15 \\
 & & unknown/declined & 10 \\
 & Race & Black or African American & 15 \\
 & & other/unknown/declined & 10 \\
\bottomrule
\caption{Complete scoring rubric across all eight point-based aggregated
feature categories. Within each category, component points are summed and
the total is capped at 100.}
\label{tab:full_rubric}
\end{longtable}

\begin{figure}[htbp]
\begin{lstlisting}[style=jsonstyle]
{
  "disease_severity": {
    "ef_points": [
      {"condition": "ef_midpoint < 30", "points": 35},
      {"condition": "30 <= ef_midpoint <= 40", "points": 25}
    ],
    "severity_cutoffs": {"medium_min": 25, "high_min": 55}
  },
  "cardiovascular_risk": {
    "continuous_components": {
      "age": {"minimum_age": 35, "maximum_age": 95, "maximum_points": 45}
    }
  },
  "medical_grounding": {
    "references": [
      {"journal": "JACC: Heart Failure", "year": 2024,
       "doi": "10.1016/j.jchf.2024.02.020"}
    ]
  }
}
\end{lstlisting}
\caption{Representative slice of the versioned scoring rubric.}
\label{fig:rubric}
\end{figure}
\endgroup

\FloatBarrier
\subsection{Stage 1 Preprocessing: Source Tables, Deduplication, Aggregation, and Signal Engineering}
\label{app:ehr_data}

Table~\ref{tab:ehr_data} lists the nine EHR-exported source tables described in Section~\ref{sec:data_task}, along with example fields from each.

\begin{table}[htbp]
\floatconts
{tab:ehr_data}
{\caption{EHR-exported source tables}}
{%
\footnotesize
\setlength{\tabcolsep}{8pt}
\renewcommand{\arraystretch}{1.4}
\begin{tabular}{@{}
>{\raggedright\arraybackslash}p{0.22\linewidth}
>{\raggedright\arraybackslash}p{0.24\linewidth}
>{\raggedright\arraybackslash}p{0.42\linewidth}
@{}}
\toprule
\textbf{Source table} & \textbf{Description} & \textbf{Example fields} \\
\midrule
PatientBase &
Demographics and core diagnoses &
Patient ID, Age, Sex, Race, Ethnicity, Language, BMI, Blood pressure (systolic/diastolic), Pulse, Primary ICD-10 code list, Primary diagnosis name (free text) \\

OtherDiagnosis &
Additional ICD diagnoses &
ICD-10 code, diagnosis name, diagnosis date \\

Medications &
Longitudinal medication history &
Medication display name, Pharmacologic class, Therapeutic class, Order date \\

EchoProcs &
Echocardiography procedures &
Procedure code, Procedure name, Order date \\

Surgery &
Surgical procedures &
Procedure name, Procedure date, Hospital, Department \\

LabsComponents &
Internal laboratory data &
Laboratory component name, Results date, Units, Abnormal flag, Procedure name \\

LabsExternal &
External laboratory data &
Test name, Result value, Units, Abnormal flag, Collection time \\

EjectionFraction &
Ejection fraction recordings &
EF value/range (e.g., ``25--30\%''), Recorded time \\

SocialHx &
Tobacco, alcohol, illicit drug history &
Tobacco use, Alcohol use, Illicit drug use \\

\bottomrule
\end{tabular}}
\end{table}

Table~\ref{tab:sources} lists the deduplication key used for each source table, described in Section~\ref{sec:standardization}.

\begin{table}[htbp]
\floatconts
{tab:sources}
{\caption{Deduplication keys used for EHR source tables.}}
{%
\footnotesize
\setlength{\tabcolsep}{9pt}
\renewcommand{\arraystretch}{1.3}
\begin{tabular}{@{}p{0.25\linewidth}p{0.60\linewidth}@{}}
\toprule
\textbf{Source table} & \textbf{Deduplication key} \\
\midrule
PatientBase & one row per \texttt{UNIQUEPATID} (enforced) \\
OtherDiagnosis & \texttt{UNIQUEPATID} + ICD list + DX name \\
Medications & \texttt{UNIQUEPATID} + \texttt{ORDERINST} + \texttt{MEDICATION\_ID} \\
EchoProcs & \texttt{UNIQUEPATID} + \texttt{ENCID} + \texttt{ORDER\_PROC\_ID} \\
Surgery & \texttt{UNIQUEPATID} + \texttt{ENCID} + \texttt{OR\_PROC\_ID} \\
Labs (internal + external) & \texttt{UNIQUEPATID} + result time + component + value \\
EjectionFraction & \texttt{UNIQUEPATID} + \texttt{RECORDED\_TIME} + entry \\
SocialHx & one row per \texttt{UNIQUEPATID} \\
\bottomrule
\end{tabular}
}
\end{table}

Table~\ref{tab:aggregates} lists the patient-level quantities aggregated from each event source before merging, described in Section~\ref{sec:method}.

\begin{table}[htbp]
\floatconts
{tab:aggregates}
{\caption{Patient-level quantities aggregated from each event source before merging.}}
{%
\footnotesize
\setlength{\tabcolsep}{5pt}
\renewcommand{\arraystretch}{1.3}
\begin{tabular}{@{}p{0.28\linewidth}p{0.62\linewidth}@{}}
\toprule
\textbf{Source domain} & \textbf{Patient-level aggregates} \\
\midrule
Medications &
Order and distinct-drug counts; first and latest order timestamps; sixteen
cardiovascular drug-class flags (e.g., beta blockers, ACE inhibitors, ARBs,
ARNI, MRAs, SGLT2 inhibitors, loop diuretics, anticoagulants, inotropes);
and derived RAAS and GDMT class counts \\
Laboratory (internal + external) &
Per-analyte counts, abnormal-result counts, maxima, and latest values, with
dedicated handling for BNP, NT-proBNP, troponin, electrolytes, creatinine,
glucose, and hemoglobin \\
Surgery &
Procedure count and procedure-type indicators (e.g., LVAD, device, valve,
catheterization, PCI) \\
Echocardiography &
Procedure count and first/latest order timestamps \\
\bottomrule
\end{tabular}
}
\end{table}

Table~\ref{tab:signals} lists the clinical-reasoning layer that converts raw text and codes into interpretable flags, described in Section~\ref{sec:method}.

\begin{table}[htbp]
\floatconts
  {tab:signals}
  {\caption{Clinical-reasoning layer mapping raw EHR signals to
  interpretable patient-level features.}}
  {%
  \footnotesize
  \begin{tabular}{@{}p{0.2\linewidth}p{0.25\linewidth}p{0.46\linewidth}@{}}
  \toprule
  \textbf{Signal} & \textbf{Source} & \textbf{Engineered representation} \\
  \midrule
  Cardiovascular comorbidities &
  ICD-10 codes and diagnosis names &
  Binary flags from curated pattern sets (e.g., heart failure, coronary artery
  disease, atrial fibrillation, chronic kidney disease, diabetes, hypertension,
  COPD) \\

  Behavioral risk &
  Social history &
  Ordered categories for tobacco and alcohol use (e.g., current vs.\ former
  tobacco use; daily, occasional, past, or no alcohol use) \\

  Biomarkers and advanced therapies &
  Laboratory results and medications &
  High-risk biomarker flags (BNP, NT-proBNP, troponin) and advanced-HF therapy
  flags (e.g., ARNI, SGLT2 inhibitors, inotropes, beta blockers) \\

  Heart-failure phenotype &
  Diagnosis names and ICD codes &
  Systolic, diastolic, or combined phenotype with a diagnosis-name/ICD
  conflict flag \\

  \bottomrule
  \end{tabular}
  }
\end{table}

\FloatBarrier
\subsection{XGBoost Hyperparameter Configuration}
\label{app:hyperparameters}
The configuration in Table~\ref{tab:xgboost_hyperparameters} emphasizes regularization through shallow trees, conservative split criteria, and row and column subsampling to reduce overfitting on the relatively small training set. The same configuration was used for all experiments so that performance differences reflected the feature sets rather than model tuning.

\begin{table}[htbp]
\floatconts
  {tab:xgboost_hyperparameters}%
  {\caption{Fixed XGBoost hyperparameter configuration used for all downstream classification experiments.}}%
  {\footnotesize
  \begin{tabular}{ll}
  \toprule
  \textbf{Hyperparameter} & \textbf{Value} \\
  \midrule
  \texttt{max\_depth} & 3 \\
  \texttt{n\_estimators} & 100 \\
  \texttt{learning\_rate} & 0.05 \\
  \texttt{gamma} & 1.0 \\
  \texttt{min\_child\_weight} & 3 \\
  \texttt{subsample} & 0.8 \\
  \texttt{colsample\_bytree} & 0.8 \\
  \bottomrule
  \end{tabular}}
\end{table}

\FloatBarrier
\subsection{LLM Rubric-Auditor Prompt}
\label{app:auditor_prompt}

The rubric-auditor step (Section~\ref{sec:method}) uses \texttt{Qwen 2.5-1.5B-Instruct} to validate each feature row against patient evidence and correct only whitelisted fields (Table~\ref{tab:llm_fields}). The abbreviated prompt is shown in Figure~\ref{fig:auditor_prompt}.

\begin{table}[htbp]
\floatconts
  {tab:llm_fields}%
  {\caption{Whitelisted fields the LLM auditor may correct, versus fields protected from modification.}}%
  {\footnotesize
  \setlength{\tabcolsep}{7pt}
  \renewcommand{\arraystretch}{0.8}
  \begin{tabular}{@{}p{0.46\linewidth}p{0.46\linewidth}@{}}
  \toprule
  \textbf{LLM-correctable} & \textbf{Protected (never modified)} \\
  \midrule
  Numeric scores, counts, indices, frequencies &
  Evidence traces (\texttt{*\_json}) \\
  Indicator / availability flags &
  Narrative rationales (\texttt{*\_explanation}) \\
  Fixed set of classification and urgency labels &
  Patient identifier (\texttt{UNIQUEPATID}) \\
  \bottomrule
  \end{tabular}}
\end{table}

\begin{figure}[htbp]
\begin{lstlisting}[style=jsonstyle]
You are a rubric auditor. Review the candidate feature row against the patient evidence.

Rules:
1. Use ONLY the patient data.
2. Use ONLY the rubric logic.
3. Do NOT recreate the entire row.
4. If every field is correct return {"valid": true}.
5. If any field is incorrect return
   {"valid": false, "corrections": {"field_name": corrected_value}}.
6. Include ONLY fields that require correction.
7. Never return the entire candidate row.
8. NEVER include any field whose name ends with "_json" or "_explanation".
9. Corrections may only contain scores, counts,
   classifications, urgency labels, indicator flags.
10-11. Do not return explanations or audit JSON.
12. Correction keys must be selected from
    correctable_fields only.

JSON only.

INPUT:
{"patient": {...}, "candidate": {...},
 "correctable_fields": [...]}
\end{lstlisting}
\caption{Abbreviated rubric-auditor prompt}
\label{fig:auditor_prompt}
\end{figure}

\onecolumn
\subsection{Comorbid Condition Discovery: Prompt and Example Output}
\label{app:comorbid_condition}

The comorbid condition discovery stage described in Section~\ref{sec:method} prompts the language model to select supporting evidence columns for each of the thirteen fixed candidate conditions. Figure~\ref{fig:condition_prompt} shows the abbreviated prompt template for one candidate condition, and Figure~\ref{fig:comorbid_condition_example} shows the representative anemia-candidate output discussed in Section~\ref{sec:results}.

\begin{figure}[htbp]
\begin{lstlisting}[style=jsonstyle]
You are creating one row for a condition-level
modeling catalog.

Use only this rubric payload and the listed final
dataset columns.
Do not assign diagnoses to individual patients.
Do not invent condition names, columns, papers,
thresholds, or rules.
Return one JSON object only:

{
  "condition_name": "exact candidate condition name",
  "prediction_columns": ["column_a", "column_b"],
  "explanation": "why these columns support
                   prediction/phenotyping/research
                   flagging"
}

RUBRIC PAYLOAD:
{"rubric_version": "2026-06-16",
 "candidate_condition": {"condition_name":
   "Anemia or hemoglobin abnormality", ...},
 "column_summary": {...}}
\end{lstlisting}
\caption{Abbreviated condition-discovery prompt template for one candidate
condition.}
\label{fig:condition_prompt}
\end{figure}

\begin{figure}[htbp]
\begin{lstlisting}[style=jsonstyle]
{
  "condition_name": "Anemia or hemoglobin abnormality",
  "condition_group": "blood",
  "evidence_type": "direct_diagnosis_plus_lab_support",
  "prediction_columns": [
    "dx_anemia", "lab_hemoglobin_count",
    "lab_hemoglobin_abnormal_count", "lab_hemoglobin_latest_value",
    "dx_chronic_kidney_disease", "feature_blood_comorbid",
    "feature_blood_comorbid_score", "feature_kidney_comorbid_score"
  ],
  "explanation": "These columns provide direct and supportive evidence
    for the Anemia or hemoglobin abnormality condition. The 'dx_anemia'
    column indicates whether anemia has been diagnosed, while
    'lab_hemoglobin_count', 'lab_hemoglobin_abnormal_count', and
    'lab_hemoglobin_latest_value' capture hemoglobin levels and their
    variability. [...]"
}
\end{lstlisting}
\caption{Representative comorbid condition-discovery output for the anemia candidate condition.}
\label{fig:comorbid_condition_example}
\end{figure}

\FloatBarrier
\subsection{LLM Construct-Validity Evaluation}
\label{app:llm_construct_validity}

The rubric was developed by clinician and methodology experts based on established prediction-model assessment standards \citep{collins2024tripodai,wolff2019probast,kahn2016dataquality,whiting2011quadas2,vasey2022decideai,wasson1985clinical,mcginn2000users}. These frameworks were synthesized into five independent dimensions: (1) external validation evidence, (2) predictor construction risk, (3) definition and reporting transparency, (4) EHR source data quality, and (5) evidence population and outcome validity. Table~\ref{tab:llm_rubric} reproduces the complete rubric.

Evaluation was performed using Claude Opus 4.8 as an isolated background agent. The model received only the grading rubric and pipeline metadata for each engineered component and was instructed to conduct live web searches for literature supporting each dimension. Unverifiable references were scored as ``no verified source found'' rather than inferred from model knowledge. Each component received a maximum of 13 points across the five dimensions. Scores were summed within categories and normalized to 0--100 by dividing by the category maximum, $13 \times$ the number of components, enabling comparison across categories of different sizes.

\begin{table*}[htbp]
\floatconts
  {tab:llm_rubric}%
  {\caption{Five-dimensional rubric used for the LLM-based evaluation of engineered clinical features.}}%
  {\begin{tabular}{
      >{\raggedright\arraybackslash}p{0.2\linewidth}
      >{\centering\arraybackslash}p{0.08\linewidth}
      >{\raggedright\arraybackslash}p{0.62\linewidth}}
  \toprule
  \textbf{Dimension} & \textbf{Points} & \textbf{Evaluation Criteria} \\
  \midrule

  External Validation Evidence &
  0--3 &
  Assesses whether the predictor, or a clinically similar proxy, appears in an externally validated multivariable prediction model (e.g., MAGGIC, SHFM, PREVENT, KFRE). Higher scores indicate stronger evidence that the predictor has previously demonstrated prognostic value in validated clinical prediction models. \\

  Predictor Construction Risk &
  0--3 &
  Evaluates whether the predictor is appropriate for prediction according to PROBAST-inspired principles, including consistent measurement, availability at prediction time, and absence of label leakage. Predictors derived directly or indirectly from outcome-defining variables receive a score of zero. \\

  Definition and Reporting Transparency &
  0--2 &
  Assesses whether the predictor can be reproduced from the documented implementation, including source variables, computation logic, timing, and handling of missing data. \\

  EHR Source Data Quality &
  0--3 &
  Evaluates the quality of the underlying structured EHR data using the Kahn data quality framework, assessing conformance to expected formats, completeness of source fields, and clinical plausibility of recorded values. \\

  Evidence Population and Outcome Validity &
  0--2 &
  Assesses whether supporting literature was conducted in clinically comparable populations using appropriate and reliable outcome definitions or reference standards. Higher scores indicate stronger applicability of the supporting evidence to the intended heart-failure population. \\

  \midrule
  \textbf{Maximum Total Score} & \textbf{13} & Sum of all five evaluation dimensions for each engineered component. Category scores are reported as normalized percentages to account for differing numbers of components per category. \\
  \bottomrule
  \end{tabular}}
\end{table*}

To illustrate the evaluation procedure, an excerpt of the LLM system prompt is provided below through Dimension~1 (External Validation Evidence). The excerpt shows the instructions governing feature-level grading, evidence verification, and the first rubric dimension. The complete evaluation protocol consisted of five dimensions as described in the main text.

\begin{systemprompt}

\textbf{SYSTEM}

Grade engineered heart-failure pipeline features across five independent
dimensions. Every feature belongs to exactly one of 11 categories:
\texttt{behavioral\_risk}, \texttt{disease\_severity},
\texttt{cardiovascular\_risk}, \texttt{kidney\_comorbid},
\texttt{lungs\_comorbid}, \texttt{brain\_comorbid},
\texttt{metabolic\_comorbid}, \texttt{blood\_comorbid},
\texttt{demographic\_vulnerability}, \texttt{gap}, and \texttt{care}.

Each category lists a ``components:'' line in PIPELINE METADATA naming its
individual weighted findings, gap checks, or derived features. The
\texttt{gap} category consolidates every feature with ``gap'' or
``gap\_count'' in its name, including
\texttt{feature\_lab\_gap\_count},
\texttt{medication\_gap\_count},
\texttt{surgery\_gap\_count},
\texttt{care\_gap\_count},
\texttt{behavior\_aware\_care\_gap},
\texttt{disease\_severity\_adjusted\_gap},
\texttt{cv\_risk\_adjusted\_gap}, and
\texttt{demographic\_vulnerability\_adjusted\_gap}.

These features are grouped together because they are either sums over,
or single-formula derivatives of, the same underlying gap checks.
The \texttt{care} category contains only
\texttt{feature\_care\_recommendation\_urgency}, which is kept separate
from \texttt{gap} because it is not itself a gap check or gap-count
derivative. No feature is graded standalone outside a category.

\medskip

\textbf{Component-level grading.}

Grade \textbf{EACH} listed component separately (0--13, across the same
five dimensions), using the category's shared rubric, citations,
implementation notes, and statistics as context. Grade each component's
citation, leakage, and transparency independently; do not simply copy
one grade across a category.

After grading a category's components, compute the rollup:

\[
\texttt{raw\_total}
=
\sum \text{component total\_points}
\]

\[
\texttt{max\_possible}
=
13 \times \text{number of components}
\]

\[
\texttt{normalized\_score}
=
\operatorname{round}
\left(
100 \times
\frac{\texttt{raw\_total}}
{\texttt{max\_possible}},
1
\right)
\]

\medskip

\textbf{The five dimensions are derived from the following methodological
frameworks:}

\begin{enumerate}
    \item \textbf{External Validation Evidence:} adapted from
    TRIPOD/PROBAST principles for predictor justification and external
    validation.

    \item \textbf{Predictor Construction Risk:} adapted from label
    leakage and prediction-timing guidance.

    \item \textbf{Definition and Reporting Transparency:} adapted from
    TRIPOD reporting recommendations.

    \item \textbf{EHR Feasibility:} adapted from common EHR
    feature-engineering and computability principles.

    \item \textbf{Clinical Plausibility:} adapted from evidence-based
    clinical prediction-model development and guideline-supported
    predictors.
\end{enumerate}

For dimensions 1 and 5, every non-zero score must be supported by a
verified source (author, year, journal, PMID/DOI if available). If no
qualifying source is found, assign 0 (or the lowest applicable band) and
state ``no verified source found.'' Never invent citations. For both
dimensions, explicitly state whether a literature search was performed.

For pipeline-specific dimensions, use only the provided PIPELINE
METADATA. Do not infer undocumented behavior.

\medskip

\textbf{Read the entire PIPELINE METADATA before grading.} Some metadata
blocks document multiple features because they share implementation
details; however, each named feature must be graded independently.
Shared documentation does not imply identical scores, particularly for
Dimension 2, where trigger conditions may differ.

Grade each feature independently without referencing scores assigned to
other features. Do not normalize or adjust scores across different
features. Process features in metadata order and return exactly 19
output objects in that order.

\medskip
\hrule
\medskip

\textbf{DIMENSION 1 -- External Validation Evidence [0--3]}

Evaluate whether this predictor, or a close clinical proxy, appears in an
externally validated, multivariable-derived cardiovascular or
heart-failure prediction model.

\begin{description}
    \item[0 =] No externally validated multivariable-derived composite
    score for heart failure or cardiovascular risk includes this
    predictor or a close clinical proxy.

    \item[1 =] Appears only in guideline-level evidence, association
    studies, or single-variable evidence -- not in a multivariable-derived
    score.

    \item[2 =] Appears as a variable in an externally validated
    multivariable-derived score (e.g., MAGGIC, SHFM, Framingham,
    ASCVD/PREVENT, KFRE), but concordance with this pipeline's use cannot
    be confirmed.

    \item[3 =] Appears in such a score and the direction and clinical
    importance of the predictor are broadly concordant with how this
    pipeline uses and weights the feature. Exact numerical equivalence
    between published coefficients and pipeline rubric points is not
    required.
\end{description}

\textbf{Required output:} named score(s), full citation, matched variable
name, citation confidence (\texttt{verified} |
\texttt{uncertain} | \texttt{not\_found}), and whether a literature
search was performed.
...
\textit{[Prompt excerpt ends here.]}
\end{systemprompt}

The model additionally received structured pipeline metadata for each
engineered component. An example of the metadata supplied for one feature
is shown below.

\begin{systemprompt}
\textbf{feature\_behavior\_risk\_score}

\textbf{category:} \texttt{behavioral\_risk}

\textbf{rubric:} \texttt{tobacco\_current=25} (else
\texttt{tobacco\_former=10}); \texttt{alcohol\_daily=15} (else
\texttt{alcohol\_current=3}, else \texttt{alcohol\_past=1});
\texttt{illicit\_drug\_current=25}; \texttt{age>=75 +5};
\texttt{ef\_latest\_midpoint\_min in (0,40) +8}; and
\texttt{dx\_chronic\_kidney\_disease}, \texttt{dx\_diabetes}, and
\texttt{dx\_coronary\_artery\_disease} each contribute +3.

\textbf{components:} \texttt{tobacco\_current},
\texttt{tobacco\_former}, \texttt{alcohol\_daily},
\texttt{alcohol\_current\_non\_daily}, \texttt{alcohol\_past},
\texttt{illicit\_drug\_current}, \texttt{age\_75\_or\_older\_modifier},
\texttt{ef\_below\_40\_modifier}, \texttt{ckd\_modifier},
\texttt{diabetes\_modifier}, \texttt{cad\_modifier}.

\textbf{clinical rationale:} Estimates lifestyle-related cardiovascular
risk, with direct behavioral evidence weighted highest and age, EF, and
comorbidity treated as smaller cross-system modifiers.

\textbf{implementation notes:} Tobacco and alcohol variables use
hierarchical precedence rules. The temporal EF definition uses the most
recent recorded EF rather than the first-recorded or lowest-ever EF.

\textbf{stats:} missing = 0.0\%, range = 0--62, mean = 8.3,
median = 6.0.
\end{systemprompt}

Table~\ref{tab:llm_construct_validity_results} reports the resulting category-level construct-validity scores, summarized in Section~\ref{sec:results}.

\begin{table}[htbp]
\floatconts
  {tab:llm_construct_validity_results}%
  {\caption{LLM-based construct-validity evaluation of engineered clinical feature categories. Scores represent summed rubric points across components and normalized category performance relative to the maximum possible score.}}%
   {\begin{tabular}{
      >{\raggedright\arraybackslash}p{0.55\linewidth}
      >{\centering\arraybackslash}p{0.25\linewidth}}
  \toprule
  \textbf{Feature Category} &
  \textbf{Normalized Score (\%)} \\
  \midrule

  Blood Comorbidities & 93.8 \\
  Cardiovascular Risk & 93.4 \\
  Brain Comorbidities & 91.3 \\
  Kidney Comorbidities & 90.6 \\
  Behavioral Risk & 89.5 \\
  Lung Comorbidities & 88.5 \\
  Metabolic Comorbidities & 85.5 \\
  Disease Severity & 74.6 \\
  Demographic Vulnerability & 63.1 \\
  Gap Features & 60.6 \\
  Care Recommendations & 38.5 \\

  \midrule
  \textbf{Overall} &
  \textbf{81.5} \\

  \bottomrule
  \end{tabular}}
\end{table}

\FloatBarrier
\subsection{Composite Feature Output Examples}
\label{app:feature_examples}

Section~\ref{sec:results} describes the two output shapes returned by the higher-order composite features: a scored assessment with ranked drivers, and a list of documentation gaps. Figure~\ref{fig:disease_severity} shows a representative disease-severity result, and Figure~\ref{fig:care_recommendation} shows a representative care-recommendation result.

\begin{figure}[htbp]
\begin{lstlisting}[style=jsonstyle]
{
  "severity": "high",
  "severity_score": 55.0,
  "hf_classification": "HFpEF_by_diagnosis_name",
  "hf_classification_source": "DXNAMES",
  "drivers": [
    "HF phenotype inferred from diagnosis text because EF is missing",
    "Markedly elevated BNP/NT-proBNP maximum",
    "Abnormal troponin result burden",
    "CKD comorbidity increases disease complexity"
  ]
}
\end{lstlisting}
\caption{Disease severity feature output.}
\label{fig:disease_severity}
\end{figure}

\begin{figure}[htbp]
\begin{lstlisting}[style=jsonstyle]
{
  "medications": [
    {"recommendation": "review statin therapy documentation for CAD",
     "evidence_column": "med_statin"}
  ],
  "labs": [
    {"recommendation": "review potassium monitoring availability",
     "evidence_column": "lab_potassium_count"},
    {"recommendation": "review creatinine monitoring availability",
     "evidence_column": "lab_creatinine_count"},
    {"recommendation": "review sodium monitoring availability",
     "evidence_column": "lab_sodium_count"}
  ],
  "surgeries": [],
  "urgency": "high",
  "recommendation_gap_count": 4
}
\end{lstlisting}
\caption{Care recommendation feature output.}
\label{fig:care_recommendation}
\end{figure}

\FloatBarrier
\subsection{SHAP Feature Importance}
\label{app:shap}

Table~\ref{tab:shap_summary} reports the complete top-10 SHAP feature rankings for both classification tasks, summarized in Section~\ref{sec:results}.

\begin{table*}[htbp]
\floatconts
  {tab:shap_summary}%
  {\caption{Top 10 features ranked by mean absolute SHAP value for the composite-feature XGBoost models. Larger values indicate greater average contribution to model predictions.}}%
  {\footnotesize
  \begin{tabular}{llr}
  \toprule
  \textbf{Task} & \textbf{Feature} & \textbf{Mean $|$SHAP$|$} \\
  \midrule
  \multirow{10}{*}{HFrEF}
  & Demographic vulnerability adjusted gap & 0.597 \\
  & Disease severity adjusted gap & 0.456 \\
  & HF GDMT medication class count & 0.425 \\
  & Medication gap count & 0.342 \\
  & Care gap count & 0.322 \\
  & Cardiovascular risk adjusted gap & 0.307 \\
  & Behavior aware care gap & 0.229 \\
  & Other diagnosis count & 0.169 \\
  & Medication order count & 0.169 \\
  & Atrial fibrillation diagnosis & 0.141 \\
  \midrule
  \multirow{10}{*}{HFpEF}
  & Disease severity score & 0.715 \\
  & Troponin abnormal count & 0.172 \\
  & Troponin max value & 0.158 \\
  & Other diagnosis count & 0.142 \\
  & Systolic blood pressure & 0.136 \\
  & Surgery count & 0.117 \\
  & Disease severity adjusted gap & 0.116 \\
  & Weight & 0.116 \\
  & B-type natriuretic peptide latest value & 0.111 \\
  & B-type natriuretic peptide max value & 0.090 \\
  \bottomrule
  \end{tabular}}
\end{table*}

\FloatBarrier
\subsection{Claude Comparator Prompt}
\label{app:claude_prompt}

As an independent comparator, we used Claude Opus 4.8, a state-of-the-art large language model, to determine whether a general-purpose LLM could autonomously engineer clinically meaningful features directly from the same raw EHR data. The model received only the EHR schema, dataset description, and a general instruction to construct features for heart-failure prediction and risk stratification. It was not provided the proposed feature schema, scoring rubric, derived feature categories, or other project-specific implementation details. The comparator was executed once using a single prompt with web search enabled, producing feature-engineering code, documentation, and a patient-level feature matrix without manual modification. The resulting feature set was evaluated using the same baseline variables, XGBoost models, hyperparameters, cross-validation protocol, performance metrics, and construct-validity grading procedure as the proposed framework.

The following prompt was provided to Claude Opus 4.8 to independently generate a patient-level feature dataset from the raw EHR exports.

\begin{systemprompt}

\textbf{SYSTEM}

You are a data scientist independently building a patient-level feature
dataset from fragmented electronic health record (EHR) exports for a
heart-failure cohort at an academic medical center.

\medskip

\textbf{DATA}

You have nine raw relational tables, all keyed on
\texttt{UNIQUEPATID}. One patient can have multiple rows in most tables,
whereas \texttt{PatientBase} and \texttt{SocialHx} contain one row per
patient. Files are located at \texttt{[PATH]/baseline\_data/}.

\begin{itemize}
\item \textbf{PatientBase}: \texttt{UNIQUEPATID, AGE, PATIENT\_SEX, GENDER\_IDENTITY, \ldots, BMI, BP\_SYSTOLIC, BP\_DIASTOLIC, PULSE, ICD10CODES, DXNAMES}.
\item \textbf{otherDiagnosis}: \texttt{UNIQUEPATID, CURRENT\_ICD10\_LIST, DX\_NAME}.
\item \textbf{Medications}: \texttt{UNIQUEPATID, ORDERINST, MEDICATION\_ID, DISPLAY\_NAME, PHARMCLASS, THERACLASS}.
\item \textbf{EchoProcs}: \texttt{UNIQUEPATID, ENCID, ORDER\_PROC\_ID, ORDER\_INST, PROC\_CODE, PROC\_NAME}.
\item \textbf{surgery}: \texttt{UNIQUEPATID, ENCID, SURGERY\_DATE, STATUS\_C, OR\_PROC\_ID, PROC\_DISPLAY\_NAME, \ldots}.
\item \textbf{LabsComponents}: \texttt{UNIQUEPATID, ENCID, ORDER\_PROC\_ID, ORDER\_INST, PROC\_CODE, ORDER\_PROCNAME, RESULT\_TIME, COMPONENT\_NAME, ABNORMAL\_YN, ORD\_VALUE, REFERENCE\_UNIT}.
\item \textbf{LabsExternal}: Same structure as \texttt{LabsComponents}, plus \texttt{COMPONENT\_ID} and \texttt{COMMON\_NAME}.
\item \textbf{EjectionFraction}: \texttt{UNIQUEPATID, ENCID, DISP\_NAME, RECORDED\_TIME, FLOWSHEET\_ENTRY, ENTRY\_COMMENTS}. \texttt{FLOWSHEET\_ENTRY} contains the raw ejection fraction value or range (e.g., ``55'', ``45--50'', ``preserved'').
\item \textbf{SocialHx}: \texttt{UNIQUEPATID, TOBACCOUSE, ALCOHOLUSE, ILLICITDRUGUSE}.
\end{itemize}

\emph{The original prompt contained the complete column schema for all
nine tables. Repetitive column lists have been abbreviated using
\texttt{\ldots} for readability; all instructional content below is
reproduced verbatim.}

This is a real, de-identified, single-institution EHR export rather than
a curated benchmark. Expect duplicate records, inconsistent terminology,
free text, and missing observations. Absence of a record means that
information was not recorded; it does not necessarily indicate a
negative finding. Treat missing and explicitly negative findings as
distinct where possible.

\medskip

\textbf{PURPOSE}

The resulting dataset will be used for downstream heart-failure
prediction and risk-stratification tasks. Prioritize features that
plausibly capture clinically meaningful information relevant to
predicting heart-failure risk.

\medskip

\textbf{TASK}

Build a single patient-level dataset with exactly one row per
\texttt{UNIQUEPATID}.

First, consolidate the nine tables by independently determining
deduplication and aggregation strategies for diagnoses, medications,
laboratory results, procedures, ejection-fraction measurements, and
other repeated records.

Second, independently engineer additional patient-level features that
you consider clinically useful for predicting and risk-stratifying heart
failure. You may transform, combine, aggregate, or summarize any
available variables. Determine the feature definitions,
representations, thresholds, and aggregation strategies yourself.

You are intentionally \emph{not} given a predefined feature schema,
feature categories, scoring rubric, prediction target, or target feature
count. Do not attempt to reproduce any existing feature-engineering
framework.

Composite features should be included when they provide a clinically
meaningful representation that is not adequately captured by their
individual components rather than being included purely for descriptive
completeness.

Where useful, you may use web search to verify clinical definitions or
thresholds. Restrict searches to general peer-reviewed clinical and
biomedical literature.

\medskip

\textbf{REQUIREMENTS}

\begin{itemize}
\item Implement the complete workflow as runnable Python using
\texttt{pandas}.
\item Every feature must be generated deterministically from the provided
EHR data.
\item Do not invent, hallucinate, or assign patient information not
supported by the source data.
\item Do not use \texttt{UNIQUEPATID}, row order, or database artifacts
as features.
\item Missing information must not automatically be interpreted as a
negative clinical finding.
\item Prioritize clinically meaningful and interpretable features rather
than maximizing feature count.
\item Document all deduplication, aggregation, and feature-engineering
decisions.
\end{itemize}

For every derived feature, document:
\begin{itemize}
\item what it measures;
\item source table(s) and column(s);
\item how it is computed;
\item how missing data are handled;
\item any clinical assumptions or thresholds used;
\item supporting literature citation, if verified through an allowed
literature search.
\end{itemize}

\medskip

\textbf{OUTPUTS}

Produce:

\begin{enumerate}
\item \texttt{comparator\_features.csv} containing one row per
\texttt{UNIQUEPATID} with the consolidated patient-level variables and
independently engineered features.
\item \texttt{comparator\_feature\_documentation.md} documenting all
aggregation and derived-feature decisions.
\item The Python script used to generate both outputs.
\end{enumerate}

Before finishing, verify that the CSV contains exactly one row per
\texttt{UNIQUEPATID}.
\end{systemprompt}

Inspection of the comparator features showed that the LLM followed a
different feature-engineering strategy from the proposed framework. Given
only the raw EHR tables, it primarily generated features based on directly
observed clinical variables, including ICD-10 diagnosis patterns,
ejection-fraction measurements, medication use, laboratory values, and
procedure history. Examples included ICD-10-derived heart-failure
phenotype and acuity indicators, a guideline-directed medical therapy
(GDMT) medication count, an ICD-10 adaptation of the Charlson
Comorbidity Index, laboratory abnormality indicators, and healthcare
utilization measures such as echocardiography frequency. Several of these
features were clinically meaningful. However, features derived from
heart-failure diagnosis codes and ejection-fraction measurements received
lower construct-validity scores because they were closely related to the
classification labels themselves, creating predictor-construction
concerns. Additionally, utilization-based features captured patterns of
care delivery rather than independently validated markers of disease
severity. These findings demonstrate that generating clinically plausible
features alone does not guarantee features that are independent,
evidence-grounded, and suitable for downstream prediction.

\FloatBarrier
\subsection{Ablation Details}
\label{app:ablation}

Eight targeted modifications were applied independently to the full feature-engineering pipeline to assess the contribution of individual feature categories and design choices. For each ablation, synthetic features were regenerated using the modified scoring logic and evaluated using the identical XGBoost training and cross-validation pipeline as the full aggregated model. This design isolates the contribution of each feature component or scoring decision by ensuring that observed performance differences arise only from the targeted ablation.

\begin{table*}[htbp]
\floatconts
  {tab:ablation_configs}%
  {\caption{Ablation configurations evaluated in the sensitivity analysis.}}%
   {\begin{tabular}{>{\raggedright\arraybackslash}p{0.2\linewidth} >{\raggedright\arraybackslash}p{0.72
\linewidth}}
  \toprule
  \textbf{Configuration} & \textbf{Description} \\
  \midrule
  Category ablation &
  Excludes all features within a composite feature category to assess its contribution. \\

  Feature ablation &
  Excludes one feature at a time to assess individual feature contributions. \\

  Equal-weight rubric &
  Replaces clinician-derived point values with equal feature weights. \\

  Random-weight rubric &
  Replaces clinician-derived point values with randomized weights. \\

  No modifiers &
  Removes cross-system comorbidity modifiers (CKD, diabetes, CAD, and hypertension) from their respective category scores. \\

  Uniform modifiers &
  Assigns a fixed 1-point weight to all cross-system modifiers to assess the contribution of modifier presence independent of magnitude. \\

  No missingness guardrails &
  Removes availability indicators and imputes missing continuous variables to assess the effect of distinguishing unmeasured from measured-normal values. \\

  Direct evidence only &
  Restricts features to diagnoses and direct measurements, excluding supportive laboratory findings, behavioral factors, and cross-system modifiers. \\
  \bottomrule
  \end{tabular}}
\end{table*}

Because repeated-CV folds reuse the same patients under different partitions and are therefore not independent trials, significance is assessed with the corrected resampled paired $t$-test of \citet{nadeau2003inference}, which inflates the variance estimate by $(1/k + n_{\text{test}}/n_{\text{train}})$ to avoid overstating confidence from correlated fold-level measurements.

\FloatBarrier
\subsection{Ablation Results}

\begin{table*}[htbp]
\floatconts
  {tab:rubric_ablation}%
  {\caption{Rubric design ablation results relative to the full aggregated model.}}%
  {\footnotesize
  \begin{tabular}{@{}p{2.3cm}p{5.5cm}cccc@{}}
  \toprule
  \textbf{Ablation} & \textbf{Modification} &
  \multicolumn{2}{c}{\textbf{HFrEF}} &
  \multicolumn{2}{c}{\textbf{HFpEF}} \\
  & & $\Delta$AUROC & $p$ & $\Delta$AUROC & $p$ \\
  \midrule
  Equal-weight rubric &
  Replaces clinician-assigned weights with equal component weighting. &
  -0.060 & 0.0059 & -0.041 & 0.0007 \\

  Random-weight rubric &
  Replaces clinician-assigned weights with randomized weights. &
  -0.058 & 0.0070 & -0.040 & 0.0010 \\

  No modifiers &
  Removes cross-system comorbidity modifiers. &
  -0.001 & 0.748 & -0.000 & 0.779 \\

  Uniform modifiers &
  Sets all modifiers to a fixed 1-point contribution. &
  -0.000 & 0.936 & -0.001 & 0.928 \\

  No missingness guardrails &
  Removes availability indicators and imputes missing values. &
  +0.001 & 0.749 & +0.007 & 0.607 \\

  Direct evidence only &
  Removes supportive evidence and retains direct measurements only. &
  -0.065 & 0.0037 & -0.040 & 0.0003 \\

  \bottomrule
  \end{tabular}}
\end{table*}

\begin{table}[htbp]
\floatconts
  {tab:category_ablation}%
  {\caption{Category-level ablation results relative to the full aggregated model. $\Delta$AUROC represents the change in mean test AUROC relative to the full model. Negative values indicate reduced performance after category removal.}}%
  {\footnotesize
  \begin{tabular}{@{}lcc@{}}
  \toprule
  \textbf{Removed feature category} &
  \textbf{HFrEF $\Delta$AUROC} &
  \textbf{HFpEF $\Delta$AUROC} \\
  \midrule
  Care gap features              & -0.039 & -0.003 \\
  Disease severity               & -0.000 & -0.033 \\
  Brain comorbid burden          & +0.001 & -0.002 \\
  Recurrence/redundancy          & -0.001 & -0.001 \\
  Metabolic comorbid burden      & -0.001 & -0.001 \\
  Demographic vulnerability      & -0.001 & -0.000 \\
  Kidney comorbid burden         & -0.000 & -0.000 \\
  Lung comorbid burden           & -0.001 & +0.001 \\
  Blood comorbid burden          & +0.001 & +0.001 \\
  Cardiovascular risk            & +0.000 & +0.001 \\
  Behavioral risk                & -0.001 & +0.003 \\
  \bottomrule
  \end{tabular}}
\end{table}

Category-level ablation identified the clinical domains contributing most to classification performance (Table~\ref{tab:category_ablation}). For HFrEF, removing care-gap features produced the largest performance reduction ($\Delta$AUROC=-0.039), whereas other category removals had minimal effects. For HFpEF, disease severity was the dominant contributor ($\Delta$AUROC=-0.033), consistent with the importance of phenotype-specific evidence. Individual feature ablations, reported below, showed similar patterns with smaller effects, suggesting that performance gains arose from combining related clinical evidence rather than dependence on single features.

\FloatBarrier
\subsection{Individual Feature Ablation}
\label{app:feature_ablation}

\begin{table*}[htbp]
\floatconts
  {tab:hfref_feature_ablation}%
  {\caption{Individual feature ablation results for HFrEF classification.}}%
  {\footnotesize
  \begin{tabular}{@{}llc@{}}
  \toprule
  \textbf{Feature category} & \textbf{Removed feature} &
  \textbf{$\Delta$AUROC} \\
  \midrule
  Metabolic comorbid burden & feature\_metabolic\_comorbid\_score & -0.002 \\
  Behavioral risk & feature\_behavior\_risk\_score & -0.001 \\
  Demographic vulnerability & feature\_demographic\_vulnerability\_score & -0.001 \\
  Lung comorbid burden & feature\_lungs\_comorbid\_score & -0.000 \\
  Care gap & feature\_lab\_gap\_count & -0.000 \\
  Care gap & feature\_surgery\_gap\_count & -0.000 \\
  Brain comorbid burden & feature\_brain\_comorbid\_score & -0.000 \\
  Care gap & feature\_care\_gap\_count & -0.000 \\
  Disease severity & feature\_disease\_severity\_score & -0.000 \\
  Blood comorbid burden & feature\_blood\_comorbid\_score & -0.000 \\
  Care gap & feature\_medication\_gap\_count & +0.000 \\
  Care gap & feature\_behavior\_aware\_care\_gap & +0.000 \\
  Cardiovascular risk & feature\_cardiovascular\_risk\_score & +0.000 \\
  Kidney comorbid burden & feature\_kidney\_comorbid\_score & +0.000 \\
  Care gap & feature\_demographic\_vulnerability\_adjusted\_gap & +0.001 \\
  \bottomrule
  \end{tabular}}
\end{table*}

\begin{table*}[htbp]
\floatconts
  {tab:hfpef_feature_ablation}%
  {\caption{Individual feature ablation results for HFpEF classification.}}%
  {\footnotesize
  \begin{tabular}{@{}llc@{}}
  \toprule
  \textbf{Feature category} & \textbf{Removed feature} &
  \textbf{$\Delta$AUROC} \\
  \midrule
  Disease severity & feature\_disease\_severity\_score & -0.033 \\
  Care gap & feature\_disease\_severity\_adjusted\_gap & -0.003 \\
  Care gap & feature\_surgery\_gap\_count & -0.002 \\
  Care gap & feature\_lab\_gap\_count & -0.001 \\
  Kidney comorbid burden & feature\_kidney\_comorbid\_high & -0.001 \\
  Care gap & feature\_medication\_gap\_count & -0.001 \\
  Kidney comorbid burden & feature\_kidney\_comorbid\_score & -0.001 \\
  Care gap & feature\_behavior\_aware\_care\_gap & -0.001 \\
  Care gap & feature\_demographic\_vulnerability\_adjusted\_gap & -0.001 \\
  Care gap & feature\_care\_gap\_count & -0.001 \\
  Metabolic comorbid burden & feature\_metabolic\_comorbid\_nan & -0.001 \\
  Kidney comorbid burden & feature\_kidney\_comorbid\_medium & -0.001 \\
  Kidney comorbid burden & feature\_kidney\_comorbid\_nan & -0.000 \\
  Lung comorbid burden & feature\_lungs\_comorbid\_low & -0.000 \\
  Lung comorbid burden & feature\_lungs\_comorbid\_nan & -0.000 \\
  \bottomrule
  \end{tabular}}
\end{table*}

\end{document}